\documentclass[runningheads]{llncs}
\usepackage{textcomp}
\usepackage{xcolor}
\usepackage{array}
\usepackage{graphicx}
\usepackage{hyperref}
\usepackage{amsmath,amssymb,amsfonts}
\usepackage[utf8]{inputenc}

\usepackage{algorithm}
\usepackage{algpseudocode}

\begin{document}
\title{Learning cooperative behaviours in adversarial multi-agent systems\thanks{This work benefited from the advice of Assistant Professor Shinkyu Park at King Abdullah University of Science and Technology (KAUST).}}

\author{
Ni Wang\inst{1}\orcidID{0000-0002-5867-0176} 
\and
Gautham P. Das\inst{2}\orcidID{0000-0001-5351-9533} 
\and
Alan G. Millard\inst{3}\orcidID{0000-0002-4424-5953}
}
\authorrunning{N. Wang, G. P. Das and A. G. Millard}
%
\institute{
Paris-Saclay University, France \\
\email{niwang.fr@gmail.com}
\and
Lincoln Agri-Robotics, University of Lincoln, United Kingdom\\
\email{gdas@lincoln.ac.uk}
\and
Department of Computer Science, University of York, United Kingdom\\
\email{alan.millard@york.ac.uk}
}
\maketitle              
\begin{abstract}
This work extends an existing virtual multi-agent platform called RoboSumo to create 
TripleSumo  --- a platform for investigating multi-agent cooperative behaviors in continuous action spaces, with physical contact in an adversarial environment. 
In this paper we investigate a scenario in which two agents, namely `Bug' and `Ant', must team up and push another agent `Spider' out of the arena.  
To tackle this goal, the newly added agent `Bug' is trained during an ongoing match between `Ant' and `Spider'. 
`Bug' must develop awareness of the other agents' actions, infer the strategy of both sides, and eventually learn an action policy to cooperate.
The reinforcement learning algorithm Deep Deterministic Policy Gradient (DDPG) is implemented with a hybrid reward structure combining dense and sparse rewards. 
The cooperative behavior is quantitatively evaluated by the mean probability of winning the match and mean number of steps needed to win.

\keywords{multi-agent cooperation  \and reinforcement learning.}
\end{abstract}
\section{Introduction}
With the success of AlphaGo, AI in games has been gaining increasing attention from researchers around the world \cite{ai}.
In the physical world, agents exhibit intelligent behaviours at multiple spatial and temporal scales \cite{human_football}.
In \cite{robosumo}, a multi-agent platform 
RoboSumo\footnote{ \href{https://github.com/openai/robosumo}{https://github.com/openai/robosumo}
} 
was designed to investigate the potential of continuous adaptation in non-stationary and competitive multi-agent environments through meta-learning. 
In that platform, two agents, either homogeneous  or heterogeneous, learn to play a `sumo' game against each other. 
The one that successfully pushes its opponent off the square arena wins the match.

While RoboSumo allows multi-agent physical interaction to be investigated in adversarial scenarios, the platform does not offer support for exploring cooperative behaviors among agents. 
In this paper, we extend RoboSumo such that a new agent is added and can team up with one of the existing agents. 
This new agent must learn a policy to cooperate with its pre-defined partner, and play against their opponent. 
We train the system with Deep Deterministic Policy Gradient (DDPG), a reinforcement learning algorithm which learns a Q-function with off-policy data and the Bellman equation, and concurrently learns a policy using the Q-function \cite{ddpg}. 
The training result is evaluated through both qualitative observations of the agents' behaviors in simulation, and two quantitative parameters -- `mean winning rate' and `steps needed to win'.
The code developed for training, testing, and evaluation is open for public access\footnote{
\href{https://github.com/niart/triplesumo}{https://github.com/niart/triplesumo}
}.
The major contributions of this work are: to establish a virtual platform that allows both cooperative and competitive interactions to be explored in physical contact-rich scenarios, and to report baseline results for the two evaluation metrics after training the system with DDPG.  

The next section of this paper reviews related work on multi-agent games based on virtual platforms; 
Section~\ref{triple} describes our extension of RoboSumo and establishes TripleSumo;
Section~\ref{experiments} details our methodology for training the agent with the DDPG algorithm, followed by an evaluation of training results. The final section summarises our findings and outlines plans for future work.
\section{Related Work}
Games provide challenging environments to quickly test new algorithms and ideas of reinforcement learning (RL) in a safe and reproducible manner \cite{google}.
Therefore, recent years have witnessed the development of a series of novel virtual game platforms that have fuelled reinforcement learning research.
While some research focuses on a single agent tackling a task, many cases are based on adversarial models, where agents compete against each other to improve the overall outcome of a system.

In 2020, `Google Research Football'\cite{google} was designed as an open-source platform where agents are trained to play football in simulated physics-based 3D scene. 
Three scenarios were provided with varying levels of difficulties.
Three RL algorithms, namely Importance Weighted Actor-Learner Architecture (IMPALA), Proximal Policy Optimization Algorithms (PPO), and Ape-X DQN, were implemented by the authors to report baselines. 
This popular platform was demonstrated to be useful in developing AI methods, for example, `TiKick' \cite{tikick}. 
However, `Google Research Football' assumes the actions are synchronously executed in multi-agent settings, limiting its utility. 
In response to this, `Fever Basketball' \cite{basketball} was developed --- an asynchronous sports game environment for multi-agent reinforcement learning.

Similarly based on virtual football games, \cite{human_football} studied integrated decision-making at multiple scales in a physically embodied multi-agent system by developing a method that combines imitation learning, single- and multi-agent reinforcement learning, and
population-based training, making use of transferable representations of behaviour.
This research evaluated agent behaviours using several analysis techniques, including statistics from real-world sports analytics.

\cite{tool} introduced a `hide-and-seek' game to investigate agents learning tool use.
Through training in this environment, agents build a series of six distinct strategies and counter strategies. 
This work suggested a promising future of multi-agent co-adaptation, which could produce complex and intelligent behaviors.

When it comes to RL-based methods for multi-agent cooperation, \cite{actor} tackled the limitation of Q-learning in a non-stationary environment, resulting in variance of the policy gradient as the number of agents grows.
They presented an adaptation of actor-critic methods that consider action policies of other agents, and a training regimen utilising an ensemble of policies for each agent that leads to more robust multi-agent policies. 

CollaQ~\cite{reward} decomposes the Q-function of each agent into a `self term' and an `interactive term', with a Multi-Agent Reward Attribution (MARA) loss that regularises the training. 
This method was validated in the `StarCraft multi-agent challenge' and was demonstrated to outperform existing state-of-the-art techniques.

In a multi-agent pursuit-evasion problem, \cite{pursuit} used shared experience to train a policy with curriculum learning for a given number of pursuers, to be executed independently by each agent at run-time. 
They designed a reward structure combining individual and group rewards to encourage good formation. 

Similarly in the pursuit-evasion problem, \cite{reducetime} presented a new geometric approach of learning cooperative behaviours in a 2-pursuer single evader scenario to reduce the capture time of the evader. 
This method was shown to be scalable thanks to categorisation and removal of redundant pursuers.

These virtual games offer useful experimental platforms and learning methods for multi-agent teamwork in adversarial environments. 
However, physical contact between interactive agents in continuous domain is rarely investigated. Our work meets this need by developing a platform that facilitates research into multi-agent cooperation in physical contact-rich environments. This paper introduces TripleSumo, which extends the RoboSumo platform, and presents reinforcement-learning of cooperative behaviours in an adversarial multi-agent setting.

\section{TripleSumo}
\label{triple} 
\begin{figure}[t]
  \centering
  \vspace{-0.5em}
  \includegraphics[width= 4in]{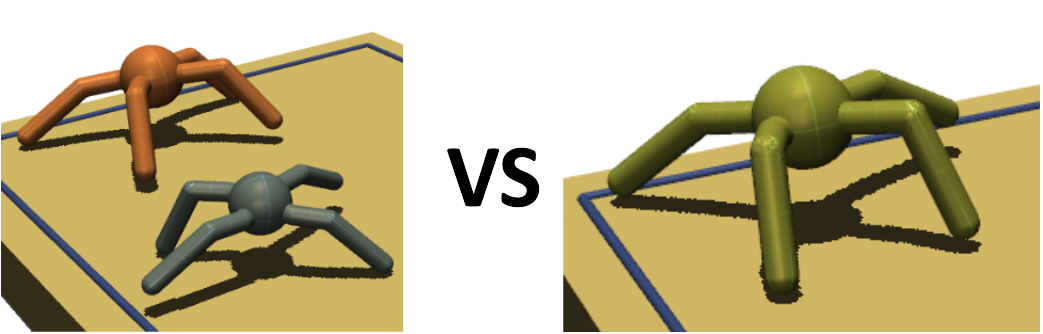}
  \caption{Competitive relation between agent `Spider' (green, right) and the pair of `Bug' (blue, middle) and `Ant' (red, left)}
  \label{vs}
  \vspace{-0.5em}
\end{figure}

Based on the RoboSumo framework built with the software toolkit OpenAI/Gym \cite{gym} and the MuJoCo\cite{mujoco} physics engine, TripleSumo adds one more agent to the system (see \figurename{~\ref{vs}}).  
\begin{figure}[htpb]
  \centering
  \includegraphics[width= 2.6in]{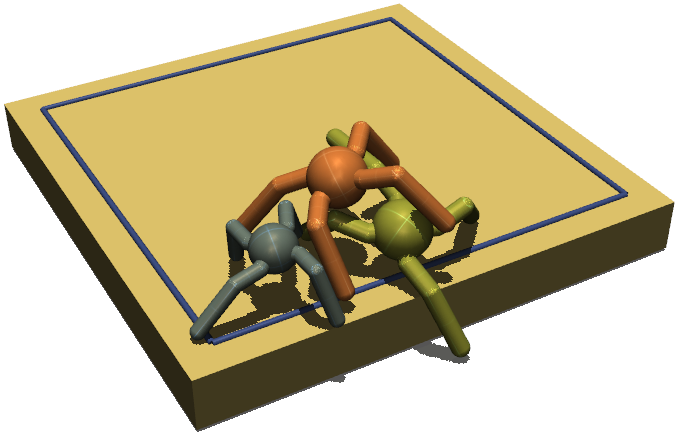}
  \caption{The simulated virtual platform `TripleSumo', where two agents (red and blue) play against the green agent on a Tatami.}
  \label{game}
  \vspace{-1.0em}
\end{figure}
In this scenario, `Ant' (red) and `Bug' (blue) team up and play against their opponent `Spider' (green) on a square arena (`Tatami') (see \figurename{~\ref{game}}). Agent behaviors are trained and observed in designated continuous action spaces.
To simplify the interfaces, this preliminary work sets up all agents to be four-legged and the same size. 
However, the three agents differ from one another in contact force (see \tableautorefname{~\ref{force}}). 
Morphological and physical features of the agents are subject to free choices according to future research demands.
Once the game starts, the three agents interact through physical contact and the match lasts until the centre of mass of any of the three agents falls outside the edge of the arena. 
The two agents `Ant' and `Spider' have been pre-trained through DDPG to create an ongoing game (see \href{https://www.youtube.com/watch?v=VVOb8t2v3pw}{supplementary video 1}).
`Spider' will win the game if it manages to push either `Ant' or `Bug' off the arena. Alternatively, the team of `Ant' and `Bug' will win the game if they manage to push off `Spider'.

\begin{table}[b]
\begin{center}
\begin{tabular}{lc} 
 \hline
& \textbf{Control range of contact force (in Newtons)}\\ [1ex] 
 \hline
`Spider' (green) & [-0.22, 0.22]\\ 

`Ant' (red) & [-0.20, 0.20]\\ 

`Bug' (blue) & [-0.18, 0.18]\\[1ex]
\hline
\end{tabular}
\end{center}
\label{force}
\caption{Control range of contact force of heterogeneous agents}
\end{table}

\section{Experimental Results}
\label{experiments}
\begin{figure}[!t]
\footnotesize
\vspace{-0.5em}
\begin{minipage}[t]{0.5\linewidth}
  \centering
  \includegraphics[width=2.7in]{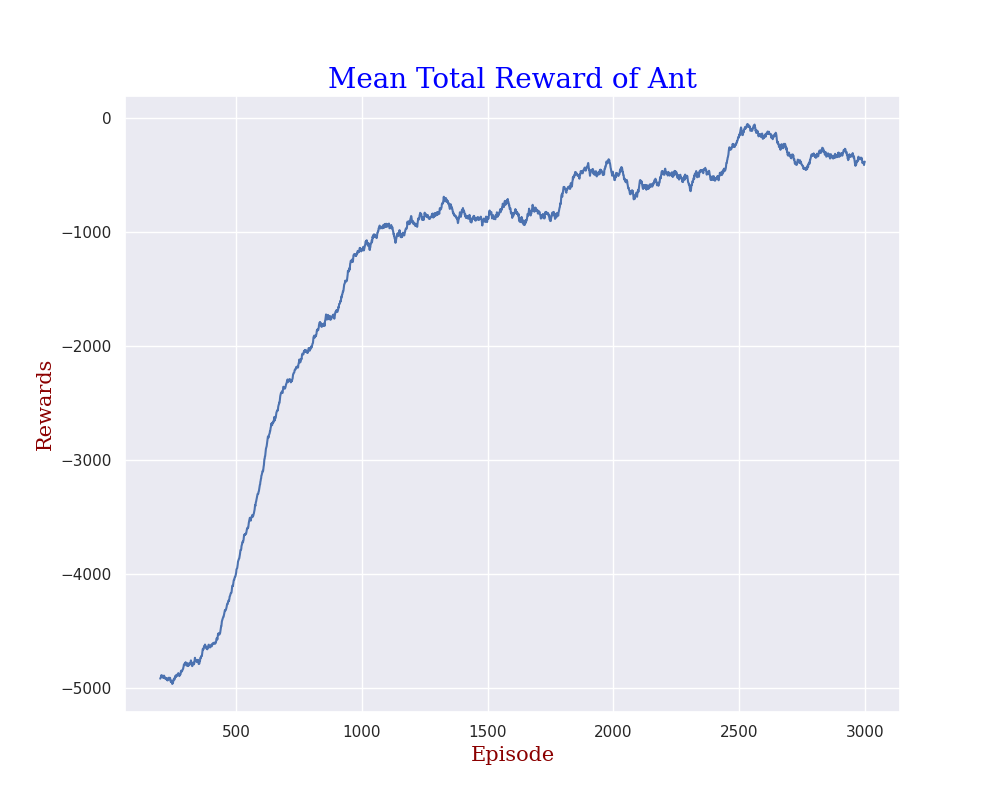}
\end{minipage}
\begin{minipage}[t]{0.5\linewidth}
\centering
\includegraphics[width=2.7in]{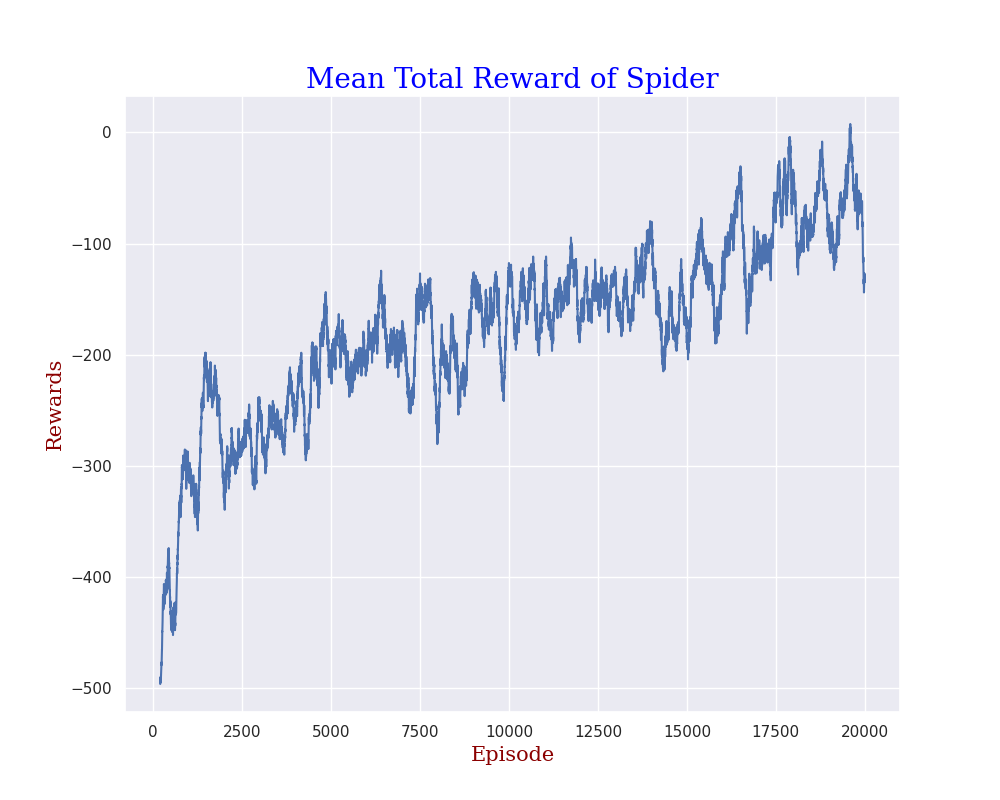}
\end{minipage}
\vspace{-2em}
  \caption{Mean total rewards received by `Ant' over 3,000 epochs of training (left); Mean total rewards received by `Spider' over 20,000 epochs of training (right); a rolling mean filter of 200 applied to each graph.}
  \label{ant_bug}
\end{figure}

This section implements a commonly used reinforcement learning algorithm DDPG to train the new agent.
In order to infer the strategies of both its opponent and partner, the agent `Bug' is trained to learn an action policy during an ongoing match\footnote{
supplementary video 1:
\href{https://www.youtube.com/watch?v=nxzi7Pha2GU}{https://www.youtube.com/watch?v=nxzi7Pha2GU}
}, where `Ant' and `Spider' are playing against each other.
The ongoing game is created by training `Ant' for 3,000 epochs with DDPG first, and afterwards training `Bug' for 20,000 epochs (\figurename~\ref{ant_bug}), with reward structures similar to equations (\ref{structure}), (\ref{dense}), but in different directions.

\subsection{Reward Shaping}
\begin{figure}[!t]
  \centering
  \includegraphics[width= 3.7in]{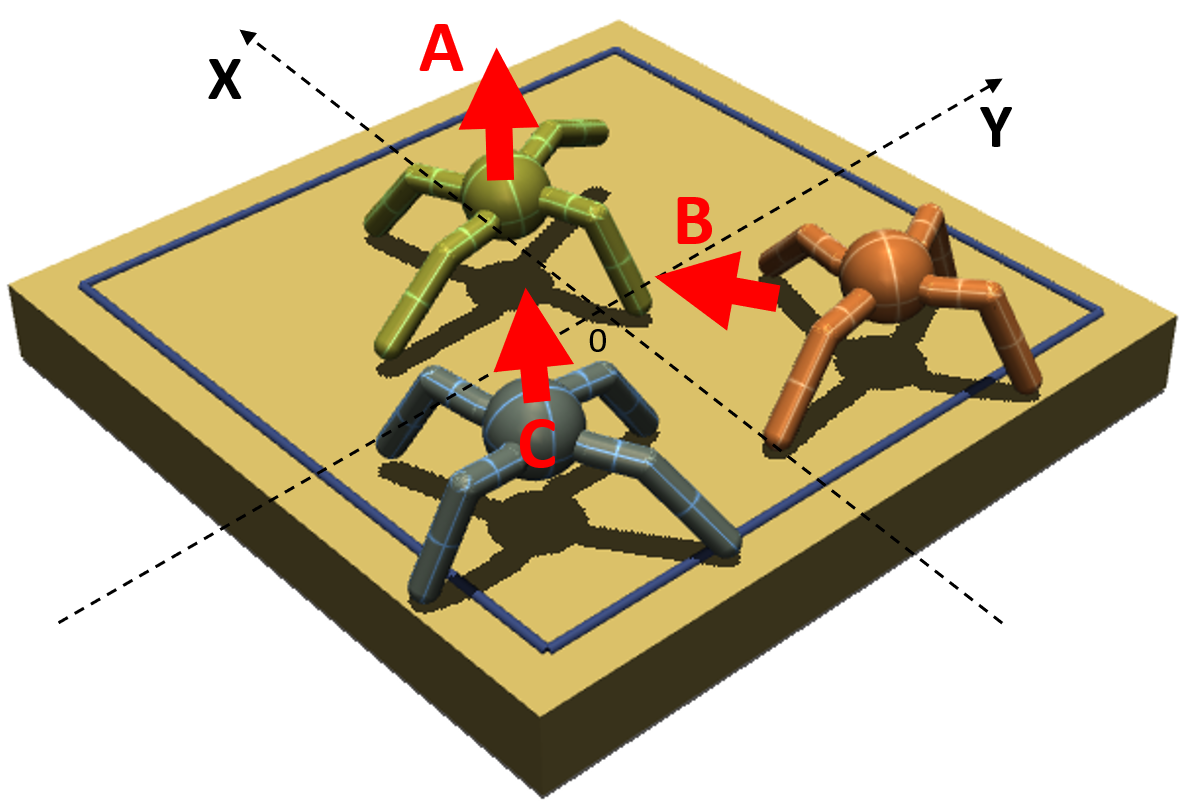}
  \vspace{-1em}
  \caption{Direction of allocating positive dense reward: A. opponent velocity reward; B. partner velocity reward; C. self velocity reward.}
  \label{direction}
\end{figure}

The reward function used in this experiment consists of two parts:
\begin{equation}
    Reward\ =  \ Dense \ Reward \ + \ Sparse \ Reward 
\label{structure}
\end{equation}
In the `Dense' part, each single step of the agent's movement is associated with a value which adds up to the total reward of the current epoch. 
This `Dense Reward' is decomposed into four terms --- `opponent velocity reward', `partner velocity reward', `self velocity reward', and a constant punishment (Equation \ref{dense}), where $C_{1}$, $C_{2}$ and $C_{3}$ are constant coefficients inserted to each term of dense reward function.
\begin{equation}
\begin{split}
    Dense \  reward = \ &C_1 \ * \ opponent \ velocity \ reward \ 
\\
+ \ &C_2 \ * \ self \ velocity \ reward \ 
\\
+ \ &C_3 \ * \ partner \ velocity \ reward \ 
\\
+ \ &still \ punishment
\end{split}
\label{dense}
\end{equation}
In each step, the agent receives an `opponent velocity reward' which is decomposed to an X component and a Y component. Its X component has an absolute value proportional to the opponent's speed along the X-direction, and is assigned to be positive if the opponent is moving backwards from the agent, and negative if the opponent moves towards the agent. 
The Y component of `opponent velocity reward' is defined in a similar way in the Y direction.
`Self velocity reward' and `partner velocity reward' have absolute values proportional to the agent's and its partner's speeds respectively, assigned to be positive if the agent or its partner moves towards its opponent, while assigned to be negative if they move away from their opponent. 
That is, the agent will be rewarded if itself or its partner moves towards their opponent, or if the opponent moves backwards from the agent, and will be punished on the contrary.
The `still punishment' is a negative constant to ensure the agent will be punished if it remains stationary.
\figurename{~\ref{direction}} shows the direction of allocating positive dense reward to the agent.
On the other hand, the `Sparse' part of the reward function is associated with the result of game. 
A score of +500 or -400 will be allocated if the team wins or loses the game respectively (see Algorithm \ref{sparse}). The sparse reward will be 0 by default if the epoch exceeds the maximum number of steps without a side winning. 

\begin{algorithm}[!b]
\caption{Sparse part of reward function}\label{alg:cap}
\begin{algorithmic}
\While{An epoch is ongoing}{
}
\If{The opponent falls off Tatami}
    \State reward += 500
\ElsIf{`Bug' or `Ant' falls off Tatami}
    \State reward -= 400
\EndIf
\EndWhile
\end{algorithmic}
\label{sparse}
\end{algorithm}
\begin{table}[t]
\begin{center}
 \begin{tabular}{p {4 cm} p {1.5 cm}} 
 \hline
\textbf{Hyper-parameter} & \textbf{Value} \\ [1ex] 
 \hline
 Optimiser & Adam\\ 

Actor Learning rate & 1e-4\\ 

Critic Learning rate & 1e-3\\ 

Discount factor ($\gamma$) & 0.99\\ 

Target Update Factor ($\tau$) & 0.01\\ 

Steps per epoch  & 500\\ 

Batch size & 64\\

Maximum epochs & 20,000 \\

Replay buffer size & 1,000,000\\
Update stride & 30\\[1ex]
\hline
\end{tabular}
\end{center}
\label{table}
\caption{Hyper-parameters in DDPG implementations}
\end{table}

\begin{figure}[htbp]
  \centering
  \vspace{-1em}
  \includegraphics[width= 4.25in]{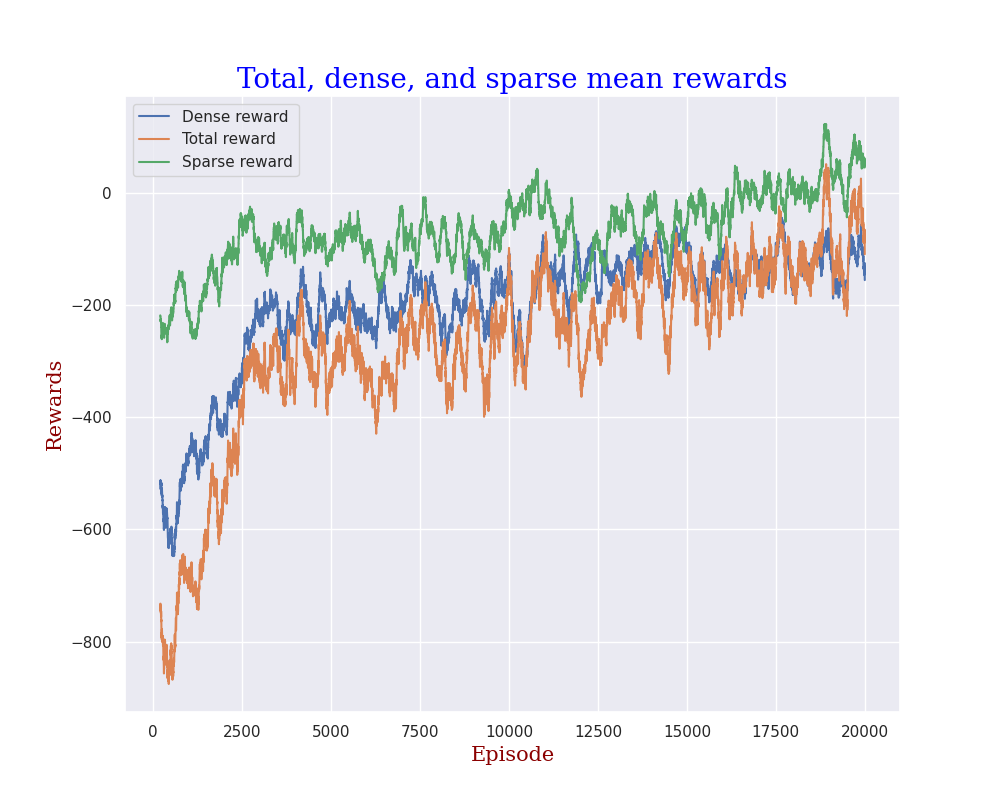}
  \vspace{-1.5em}
  \caption{Mean of Total rewards (orange), Dense reward (blue), and Sparse reward (green) received by `Bug' in each epoch during training. 
A rolling mean filter of 200 is applied.
The final dense and total rewards stay below zero due to a constant `still punishment' term in reward function.}
  \label{rewards}

  \centering
  \includegraphics[width= 4.25in]{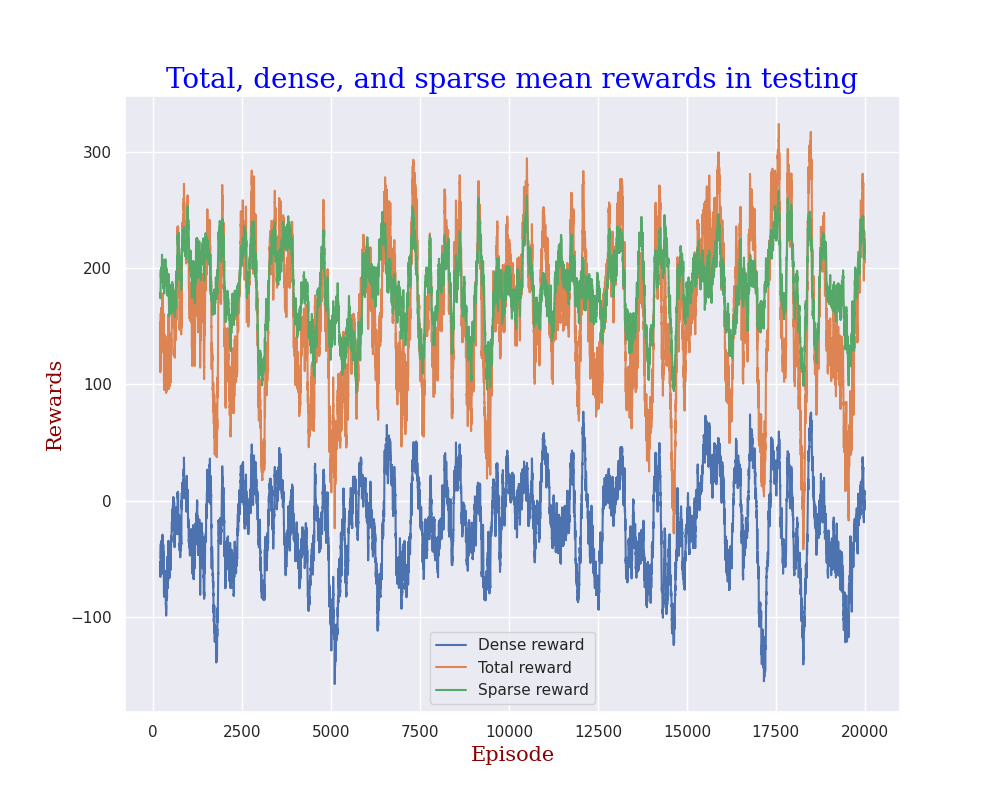}
  \vspace{-1.5em}
  \caption{Mean of Total rewards (orange), Dense reward (blue), and Sparse reward (green) received by `Bug' in a test of 20,000 epochs. 
A rolling mean filter of 200 is applied.}
  \label{testing}
\end{figure}

Important hyper-parameters are summarised in \tableautorefname{~\ref{table}}. Over the course of training for 20,000 epochs, the agent demonstrates increasing mean values of both dense and sparse rewards, as shown in \figurename{~\ref{rewards}}.

\begin{figure}[htbp]
  \vspace{-1em}
  \centering
  \includegraphics[width=4.2in]{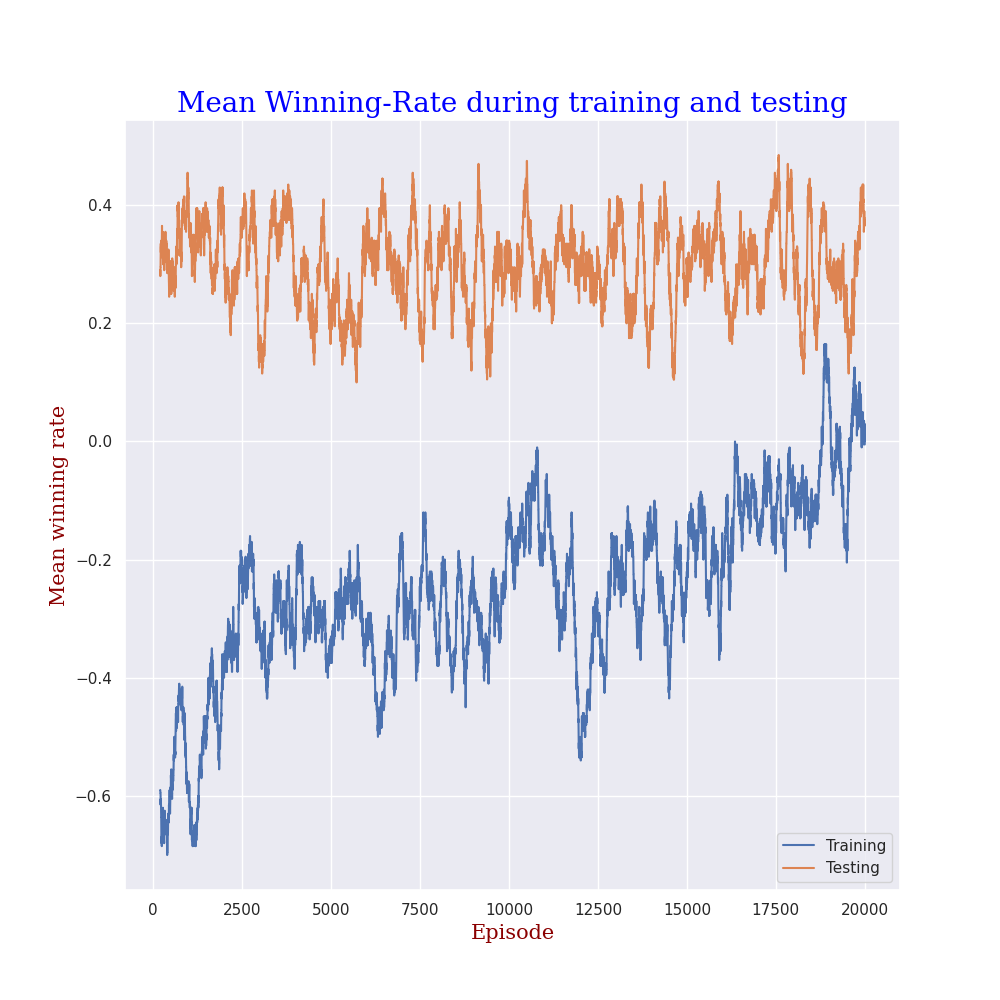}
  \vspace{-1.5em}
  \caption{
MWR of the team increases as training progresses (blue). The MWR rising from negative to positive indicates the team ends up with stronger competency than its opponent.
MWR of the team remains positive (between 0.15 and 0.47) in testing of 20,000 epochs (orange). A rolling mean filter of 200 was applied to each graph.}
  \label{rate}

  \centering
  \includegraphics[width= 4.2in]{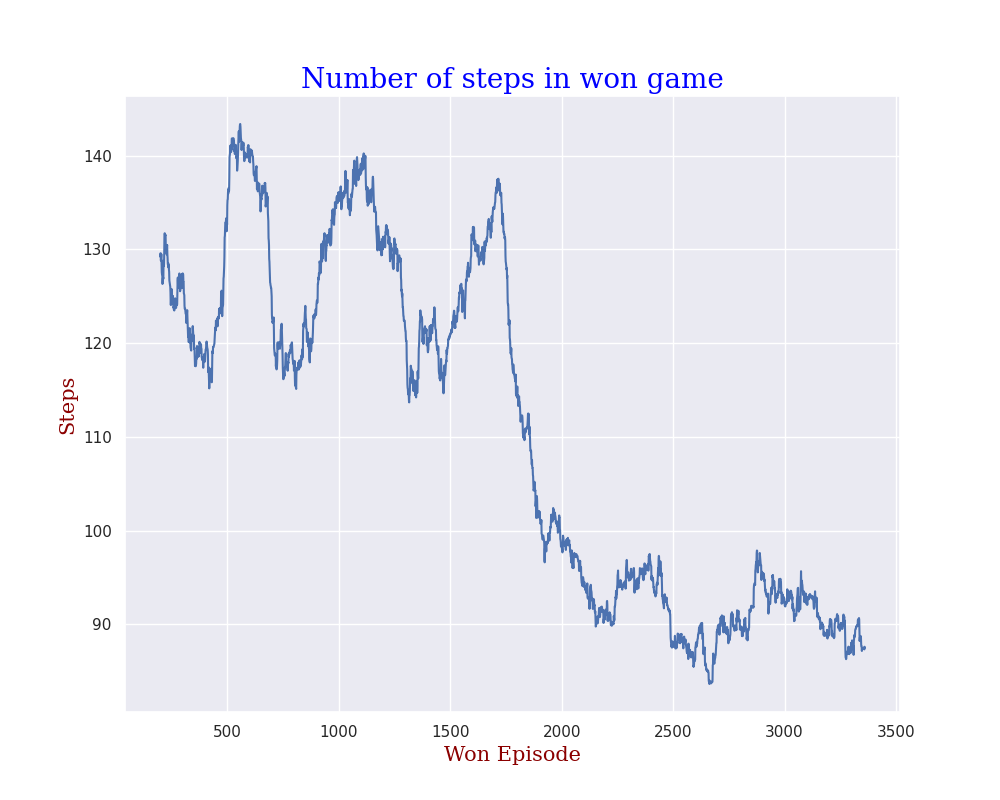}
  \vspace{-1.5em}
  \caption{Mean value of number of steps needed to win the game decreases over the course of training. 
A rolling mean filter of 200 is applied. 
Only games won by the team are counted.}
  \label{steps}
\end{figure}

\subsection{Performance Evaluation}
The action policy of `Bug' resulting from this training is observed in a test
across 20,000 epochs\footnote{
supplementary video 2:
\href{https://www.youtube.com/watch?v=C4xGuIyeY5A}{https://www.youtube.com/watch?v=C4xGuIyeY5A}
}, where the agents demonstrate cooperative behavior and stable values of both dense and sparse rewards (see \figurename{~\ref{testing}}).
Additionally, in order to evaluate the efficiency of agents' cooperation, we used the two metrics: MWR and `steps needed to win' to analyse the training result. 

\begin{table}[t]
\begin{center}
 \begin{tabular}{p {4 cm} >{\centering\arraybackslash}p{4cm}} 
 \hline
\textbf{Result of an epoch} & \textbf{Local winning rate}\\ [1ex] 
 \hline
Team wins & +1\\ 

Team loses & -1\\ 

Time expires without result & 0\\
\hline
\end{tabular}
\end{center}
\caption{Definition of local winning rate at a certain epoch}
\label{wr}
\vspace{-1em}
\end{table}

A metric `winning rate' (WR) is defined as the probability of winning the game at a certain phase during training or testing.
During training, the local WR at a certain epoch is assigned to be `+1' if the team wins the game; assigned to be `-1' if the team loses the game; and assigned to be `0' if the epoch reaches maximum number of steps without winning or losing. 
The definition of local WR is described in \tableautorefname{~\ref{wr}}.
A `mean WR' is calculated as the rolling mean value of local winning rate when the training or testing reaches a certain epoch. 
Intuitively, a positive WR indicates the team is stronger than opponent, while a negative WR indicates the team is weaker than its opponent, and a zero WR indicates the two sides have similar levels of competency. 
With a rolling mean filter of $\lambda$, the mean WR (denoted as MWR) at epoch $\alpha$ (denoted as $E_{\alpha}$) is:
\begin{equation}
    MWR_{\alpha} \ = \frac{sum \ of \ all \ WRs \ in \ recent \ \lambda \ epochs \ until \ E_{\alpha}}{\lambda}
\label{mwr}
\end{equation}
The variations of MWR during training and testing are plotted in \figurename{~\ref{rate}}, which indicate that the team's competency improves as the training progresses, and remains stable in the resulting policy. 
In other words, the agent `Bug' has learned to cooperate with `Ant' and contribute to the teamwork.

Another metric `number of steps needed to win a game' is defined to reflect the cost of time for a successful teamwork. 
Its rolling mean (computed in a way similar to Equation ~\ref{mwr}) reflects the variation of teamwork efficiency during the training process.
In \figurename{~\ref{steps}}, the plot indicates that the team reaches higher levels of efficiency in teamwork as they are trained, resulting in reducing the amount of steps or time needed to win the game.

\section{Discussion \& Future work}
This preliminary work extended an existing virtual multi-agent platform RoboSumo into TripleSumo, which contains three players in a sumo game.
Two agents are predefined to team up and play against the other agent.
As a baseline, cooperative behaviors were investigated by training the newly added agent with the reinforcement learning algorithm DDPG, using a hybrid reward structure during an ongoing match. 
Both the sparse and dense parts of mean rewards are demonstrated to increase and eventually converge as the training process progresses.
The teamwork increases in competency and efficiency, as reflected by an increasing mean winning rate and a decreasing mean number of steps needed to win a game, which indicates successful cooperation between the agents in this adversarial environment. While this work has focussed exclusively on DDPG, our future research will investigate the peformance of other RL algorithms with respect to learning cooperative strategies in multi-agent systems.

The scenario presented in this paper is similar to learning cooperative behaviours in a 2-pursuer single evader scenario to reduce the capture time of the evader, with direct applications in swarm robotic systems. A natural extension is to increase the complexity of the environment, to include more pursuing agents making it a M-Pursuer single-evader scenario and incorporating probabilistic observations about the prey positions. It would then be possible to evolve different categorical behaviours such as `interceptor', `escort', and `redundant' for the pursuer agents. 

The next step of this research is to investigate a more complex scenario of non-predefined pairing for cooperation in TripleSumo.
With three or more agents playing against one another, the game will continue until only one agent remains on the arena.
All agents will learn to freely choose to team up with a non-predefined partner in order to remain as long as possible in the game.

\bibliographystyle{IEEEtran}
\bibliography{citation}

\begin{thebibliography}{10}
\providecommand{\url}[1]{#1}
\csname url@samestyle\endcsname
\providecommand{\newblock}{\relax}
\providecommand{\bibinfo}[2]{#2}
\providecommand{\BIBentrySTDinterwordspacing}{\spaceskip=0pt\relax}
\providecommand{\BIBentryALTinterwordstretchfactor}{4}
\providecommand{\BIBentryALTinterwordspacing}{\spaceskip=\fontdimen2\font plus
\BIBentryALTinterwordstretchfactor\fontdimen3\font minus
  \fontdimen4\font\relax}
\providecommand{\BIBforeignlanguage}[2]{{%
\expandafter\ifx\csname l@#1\endcsname\relax
\typeout{** WARNING: IEEEtran.bst: No hyphenation pattern has been}%
\typeout{** loaded for the language `#1'. Using the pattern for}%
\typeout{** the default language instead.}%
\else
\language=\csname l@#1\endcsname
\fi
#2}}
\providecommand{\BIBdecl}{\relax}
\BIBdecl

\bibitem{ai}
Q.~Yin, J.~Yang, W.~Ni, B.~Liang, and K.~Huang, ``{AI in Games: Techniques,
  Challenges and Opportunities},'' \emph{ArXiv}, vol. abs/2111.07631, 2021.

\bibitem{human_football}
S.~Liu, G.~Lever, Z.~Wang, J.~Merel, S.~M.~A. Eslami, D.~Hennes, W.~M.
  Czarnecki, Y.~Tassa, S.~Omidshafiei, A.~Abdolmaleki, N.~Siegel,
  L.~Hasenclever, L.~Marris, S.~Tunyasuvunakool, H.~F. Song, M.~Wulfmeier,
  P.~Muller, T.~Haarnoja, B.~D. Tracey, K.~Tuyls, T.~Graepel, and N.~M.~O.
  Heess, ``From motor control to team play in simulated humanoid football,''
  \emph{ArXiv}, vol. abs/2105.12196, 2021.

\bibitem{robosumo}
M.~Al-Shedivat, T.~Bansal, Y.~Burda, I.~Sutskever, I.~Mordatch, and P.~Abbeel,
  ``Continuous adaptation via meta-learning in nonstationary and competitive
  environments,'' \emph{ArXiv}, vol. abs/1710.03641, 2018.

\bibitem{ddpg}
T.~P. Lillicrap, J.~J. Hunt, A.~Pritzel, N.~M.~O. Heess, T.~Erez, Y.~Tassa,
  D.~Silver, and D.~Wierstra, ``Continuous control with deep reinforcement
  learning,'' \emph{CoRR}, vol. abs/1509.02971, 2016.

\bibitem{google}
K.~Kurach, A.~Raichuk, P.~Stanczyk, M.~Zajac, O.~Bachem, L.~Espeholt,
  C.~Riquelme, D.~Vincent, M.~Michalski, O.~Bousquet, and S.~Gelly, ``Google
  research football: A novel reinforcement learning environment,'' in
  \emph{AAAI}, 2020.

\bibitem{tikick}
S.~Huang, W.~Chen, L.~Zhang, Z.~Li, F.~Zhu, D.~Ye, T.~Chen, and J.~Zhu,
  ``{TiKick: Towards Playing Multi-agent Football Full Games from Single-agent
  Demonstrations},'' \emph{ArXiv}, vol. abs/2110.04507, 2021.

\bibitem{basketball}
H.~Jia, Y.~Hu, Y.~Chen, C.~Ren, T.~Lv, C.~Fan, and C.~Zhang, ``Fever
  basketball: A complex, flexible, and asynchronized sports game environment
  for multi-agent reinforcement learning,'' \emph{ArXiv}, vol. abs/2012.03204,
  2020.

\bibitem{tool}
B.~Baker, I.~Kanitscheider, T.~Markov, Y.~Wu, G.~Powell, B.~McGrew, and
  I.~Mordatch, ``Emergent tool use from multi-agent autocurricula,''
  \emph{ArXiv}, vol. abs/1909.07528, 2020.

\bibitem{actor}
R.~Lowe, Y.~Wu, A.~Tamar, J.~Harb, P.~Abbeel, and I.~Mordatch, ``Multi-agent
  actor-critic for mixed cooperative-competitive environments,'' \emph{ArXiv},
  vol. abs/1706.02275, 2017.

\bibitem{reward}
T.~Zhang, H.~Xu, X.~Wang, Y.~Wu, K.~Keutzer, J.~Gonzalez, and Y.~Tian,
  ``Multi-agent collaboration via reward attribution decomposition,''
  \emph{ArXiv}, vol. abs/2010.08531, 2020.

\bibitem{pursuit}
C.~de~Souza, R.~Newbury, A.~Cosgun, P.~Castillo, B.~Vidolov, and D.~Kuli{\'c},
  ``Decentralized multi-agent pursuit using deep reinforcement learning,''
  \emph{IEEE Robotics and Automation Letters}, vol.~6, pp. 4552--4559, 2021.

\bibitem{reducetime}
A.~Von~Moll, D.~W. Casbeer, E.~Garcia, and D.~Milutinović, ``Pursuit-evasion
  of an evader by multiple pursuers,'' in \emph{2018 International Conference
  on Unmanned Aircraft Systems (ICUAS)}, 2018, pp. 133--142.

\bibitem{gym}
G.~Brockman, V.~Cheung, L.~Pettersson, J.~Schneider, J.~Schulman, J.~Tang, and
  W.~Zaremba, ``Openai gym,'' \emph{ArXiv}, vol. abs/1606.01540, 2016.

\bibitem{mujoco}
E.~Todorov, T.~Erez, and Y.~Tassa, ``{MuJoCo}: A physics engine for model-based
  control,'' in \emph{2012 IEEE/RSJ International Conference on Intelligent
  Robots and Systems}, 2012, pp. 5026--5033.

\end{thebibliography}
\end{document}